\documentclass[lettersize,journal]{IEEEtran}
\usepackage{amsmath,amsfonts}
\usepackage{algorithmic}
\usepackage{algorithm}
\usepackage{array}
\usepackage[caption=false,font=normalsize,labelfont=sf,textfont=sf]{subfig}
\usepackage{textcomp}
\usepackage{stfloats}
\usepackage{url}
\usepackage{verbatim}
\usepackage{graphicx}
\usepackage{cite}
\usepackage{amsfonts}  
\usepackage{algorithm}
\usepackage{algorithmic}
\usepackage{siunitx}   
\usepackage{upgreek}   
\usepackage{xcolor}    
\usepackage{booktabs}       
\usepackage{longtable}
\usepackage{threeparttable}
\usepackage{multirow}
\usepackage{multicol}
\usepackage{times}
\usepackage{hyperref}

\usepackage{amsthm}

\def\x{\mathbf{x}}

\usepackage{colortbl}
\usepackage{makecell}

\definecolor{LightCyan}{rgb}{0.6,0.6,0.5}
\definecolor{Gray}{gray}{0.85}
\definecolor{Gray2}{gray}{0.75}
\definecolor{Gray3}{gray}{0.65}
\definecolor{Gray4}{gray}{0.55}
\definecolor{Gray5}{gray}{0.45}

\def\S{\mathcal{S}}

\def\S{\mathcal{S}}

\def\x{\boldsymbol{x}}

\def\W{\mathbf{W}}
\def\w{\mathbf{w}}

\begin{document}

	
\title{\huge\bf Towards NeuroAI: Introducing Neuronal Diversity into Artificial Neural Networks}

\author{Feng-Lei Fan, \textit{Member, IEEE}, Yingxin Li, Hanchuan Peng, \textit{Fellow, IEEE}, Tieyong Zeng\textsuperscript{\rm *}, Fei Wang\textsuperscript{\rm *}
	\thanks{*Tieyong Zeng and Fei Wang are co-corresponding authors.}
	\thanks{Feng-Lei Fan and Tieyong Zeng are with Department of Mathematics, The Chinese University of Hong Kong, Hong Kong, HK 999077}
	\thanks{Ms. Yingxin Li and Hanchuan Peng are with SEU-ALLEN Joint Center, Institute for Brain and Intelligence, Southeast University, Nanjing, Jiangsu Province 210096, PR China.}
	\thanks{Dr. Fei Wang is with Department of Population Health Sciences, Weill Cornell Medicine, Cornell University, New York, NY 10065, USA.}
}

\markboth{Journal of \LaTeX\ Class Files,~Vol.~14, No.~8, Jan~2023}%
{Shell \MakeLowercase{\textit{et al.}}: A Sample Article Using IEEEtran.cls for IEEE Journals}

	
\maketitle

\begin{abstract}
Throughout history, the development of artificial intelligence, particularly artificial neural networks, has been open to and constantly inspired by the increasingly deepened understanding of the brain, such as the inspiration of neocognitron, which is the pioneering work of convolutional neural networks. Per the motives of the emerging field: NeuroAI, a great amount of neuroscience knowledge can help catalyze the next generation of AI by endowing a network with more powerful capabilities. As we know, the human brain has numerous morphologically and functionally different neurons, while artificial neural networks are almost exclusively built on a single neuron type. In the human brain, neuronal diversity is an enabling factor for all kinds of biological intelligent behaviors.   Since an artificial network is a miniature of the human brain, introducing neuronal diversity should be valuable in terms of addressing those essential problems of artificial networks such as efficiency, interpretability, and memory. In this perspective, we first discuss the preliminaries of biological neuronal diversity and the characteristics of information transmission and processing in a biological neuron. Then, we review studies of designing new neurons for artificial networks. Next, we discuss what gains can neuronal diversity bring into artificial networks and exemplary applications in several important fields. Lastly, we discuss the challenges and future directions of neuronal diversity to explore the potential of NeuroAI.

\end{abstract}


	
\section{INTRODUCTION} \label{sec:introduction}
We are experiencing the third wave of the artificial intelligence revolution. Deep learning, represented by deep artificial neural networks, has been dominating numerous important research fields in the past decade \cite{fan2021sparse,jumper2021highly,floridi2020gpt}. The tale of artificial neural networks dates back to mimicking nervous systems \cite{mcculloch1943logical} in 1943, where McCulloch and Pitts abstracted a nervous system as a net of neurons, and each neuron was modeled with a 'threshold logic' because the neuron exhibits an all-or-none behavior. Then, Hebb was engaged by the connectionism idea and proposed the rule of Hebbian learning---neurons that fire together should be wired together \cite{hebb2005organization}. The Hebbian rule now still plays an important role in neural networks and computational neuroscience to this day. Later, Rosenblatt extended McCulloch and Pitts's idea to propose the Perceptron \cite{rosenblatt1958perceptron} that computes the inner product of the input and the Perceptron's internal parameters followed by a nonlinear activation. The Perceptron is the simplest network structure whose weights and bias are automatically learned by contrasting the outputs of the Perceptron and the target. However, the Perceptron was suggested to have an extremely limited expressive ability in Minsky and Papert's book \cite{marvin1969perceptrons}, \textit{i.e.}, Perceptron cannot even execute the XOR logic. Hence, the research of Perceptron, unfortunately, entered into winter. Until 1980s, the designs of Hopfield networks \cite{hopfield1982neural} and Boltzmann machines \cite{hinton1986learning} greatly evoked the resurgence of interest in neural networks. With the re-invention of backpropagation algorithm \cite{rumelhart1986learning,werbos1994roots}, studies of neural networks were rapidly advancing: many novel network models (LSTM \cite{hochreiter1997long}, autoencoder \cite{rumelhart1985learning}, neocognitron \cite{fukushima1982neocognitron}, CNN \cite{lecun1998gradient}) were established, and finally caught the imagination of the world in 2012 by the AlexNet \cite{krizhevsky2012imagenet}'s top performance on the ImageNet \cite{deng2009imagenet}.

The brain is the most intelligent system we have ever known so far. Throughout history, the development of artificial neural networks has been open to and constantly inspired by the increasingly deepened understanding of the brain \cite{zador2022toward}. Although after drawing inspiration from neuroscience, artificial neural networks would usually follow their own paths to fit the demands of real-world tasks. In the past decades, due to tens of billions of dollars invested into neuroscience such as NIH BRAIN initiative\footnote{https://braininitiative.nih.gov/}, a great amount of knowledge regarding the brain has been amassed, which can provide an ample source for applying principles of the brain intelligence to artificial neural networks. Therefore, our opinion is that neuroscience is still critical to the advances of artificial neural networks, given the incomparable capability of the human brain and an ever-growing understanding of brain intelligence.
Our opinion well aligns with the premise of an emerging interdisciplinary field---NeuroAI \cite{zador2022toward}. The overarching goal of NeuroAI is to catalyze the next generation of AI by endowing a network with more human-like capabilities. For the time being, although a human brain and an artificial neural network serve fundamentally different purposes, it is crystal clear that the existing artificial network still goes far behind our human brain from the perspective of engineering:

\textbullet~\textbf{Efficiency:} With the advent of big models, the size of a neural network model becomes increasingly larger. To train such a model to a practical point, well-curated big data and considerable power consumption need to be supplied \cite{garcia2019estimation}. For example, the well-performed language model GPT-3 has 175 billion parameters, and its training requires hundreds of GPUs running a few months on 45 TB text data \cite{brown2020language}. In contrast, a human brain performs its incredible feat by managing billions of neurons and coordinating trillions of connections at extremely low power (<20W) \cite{cox2014neural}. This gap is because our brain is an efficient concept learner that can learn complex objects from just a few examples \cite{lake2011one}. The efficiency issue gains more and more traction in the background of combating global warming \cite{dhar2020carbon}, as the AI models, particularly big language models are widely recognized as a significant carbon emitter \cite{strubell2019energy}. Environmentalists criticized that oftentimes, the exhaustive trial-and-error fine-tuning only leads to little performance gain. According to the carbon footprint computation, the training of the BERT model has a carbon footprint close to a person's one round-trip trans-America flight\footnote{https://www.technologyreview.com/2019/06/06/239031/training-a-single-ai-model-can-emit-as-much-carbon-as-five-cars-in-their-lifetimes/}.

\textbullet~\textbf{Interpretability:} A neural network is notoriously a black box \cite{fan2021interpretability}. Although a network performs quite well in real-world tasks, it is hard to explain the underlying mechanism. Questions are often asked what is the function of certain neurons, layers, blocks, etc. and how they impact the model's decision-making. However, only limited success is achieved for these questions. Interpretability studies are divided into two branches \cite{fan2021interpretability}: \textit{post hoc} interpretation and \textit{ad hoc} interpretable modeling. It was argued that post hoc interpretation cannot be completely faithful to the original model because if it can be, it becomes the original model \cite{rudin2019stop}. What's worse is one can hardly know the nuance between the post hoc interpretation and the original model. But ad hoc interpretable modeling may suffer from model expressibility in accomplishing transparency. As opposed to deep models, the decision process in the human is highly tractable. Modern neuroscience attributes conscious acts to electrical and chemical changes within and across neurons. Visualization of working regions and neurons in both time and space can also be carried out well by modern brain imaging techniques such as fMRI \cite{heeger2002does}.

\textbullet~\textbf{Memory:} Catastrophic forgetting is a common issue of connectionist models \cite{mccloskey1989catastrophic,ratcliff1990connectionist}, \textit{i.e.}, artificial neural networks are incapable of learning new information without forgetting what is previously learned. When a network is trained to learn consecutive tasks, what is learned from the previous task is easy to be interrupted by what is learned from the current task. This is because the weights trained for the earlier task have to be changed to meet the objective of the new task. The existing solutions to overcome catastrophic forgetting either require explicit retraining using the old data (so-called interleaved learning \cite{hasselmo2017avoiding}) or
only show efficacy in a specific type of memory or network structures \cite{kirkpatrick2017overcoming}. 
Compared to the artificial network, our human brain has evolved effective mechanisms to avoid catastrophic forgetting \cite{gonzalez2020can}, \textit{e.g.}, sleeping \cite{paller2004memory} can consolidate the outcome of the awake-state learning. Although these mechanisms remain to be completely understood, more and more experiments on the neuronal levels showed that sleeping facilitates the potential of target neurons to be evoked \cite{wilson1994sleep,wilson2006awake}.

\textbullet~\textbf{Robustness:} Lacking robustness is another Achilles' heel of a neural network \cite{carlini2017towards}. It was often reported that a neural network is easy to trick \cite{tencent2019experimental}. Sometimes, adding noise that is indiscernible to a human can completely change a network's prediction \cite{szegedy2013intriguing}. Later, it was shown that a neural network can be severely interfered with by common perturbation \cite{hendrycks2019benchmarking}, such as occlusion, blur in an image, etc. Due to the widespread deployment of AI models in mission-critical scenarios, the robustness issue of neural networks has received lots of attention. Compared to a neural network, humans usually won't get confused despite small changes such as mask, shift, distortion, natural corruptions, and stylish changes in an image.

To address these issues to promote neural networks to a higher level of intelligence based on NeuroAI, we believe that at present an actionable way is to explicitly identify the differences between brain and artificial neural networks and then make efforts to mitigate these differences. Clearly, the current mainstream neural network models are remarkedly different from the biological neural system, and one primary distinction is that neural networks lack the neuronal diversity that is everywhere in the human brain \cite{stevens1998neuronal}. Different from artificial networks that are built on a single universal primitive neuron type, the brain has numerous morphologically and functionally different neurons \cite{peng2021morphological}. With no exaggeration, neuronal diversity is an enabling factor for all kinds of intelligent behaviors \cite{thivierge2008neural}. More and more works showed the biological effect of neuronal diversity. Krishnan Padmanabhan and Nathaniel N Urban examined the outputs from a single type of neuron and the mitral cells of the mouse olfactory bulb, and found that diverse populations were able to code for twofold more information than their homogeneous counterparts using this intrinsic heterogeneity \cite{urban2010diverseencode}. Since the artificial neural network is a miniature of the biological neural network, introducing neuronal diversity should be able to shed light on the aforementioned problems of the artificial neural network.

\begin{figure}
\centering
\includegraphics[width=\linewidth]{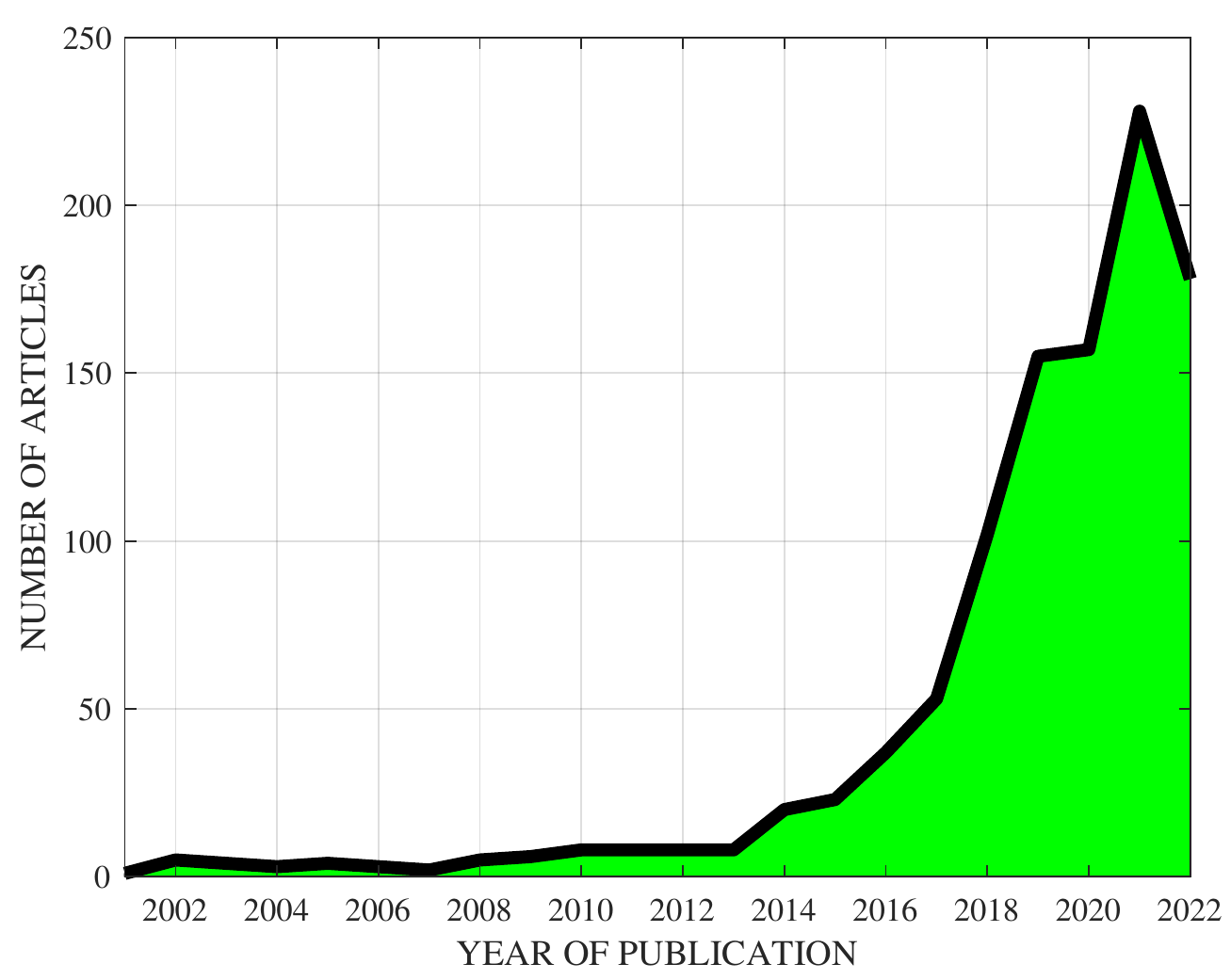}
\caption{The rapid growth of the number of articles on the research of new neurons in deep learning. The data are based on the search in the Web of Science on December 12, 2022, with the time range from 2000 to 2022 according to the keyword “Deep Learning New Neurons”.} 
\label{fig:trend}
\end{figure}

In the past years, the research of introducing neuronal diversity has attracted more and more interest, which is to design new neurons and involve them in the models. We enter the search term “Deep Learning New Neurons” into the Web of Science on December 12, 2022, with the time range from 2000 to 2022. The number of queried articles with respect to the year of publication is plotted in Figure \ref{fig:trend}, which clearly reveals an exponential trend in this topic. However, a systematic summary of introducing new neurons into deep learning, including an up-to-date review and outlook, remains overdue.

In this perspective, we first summarize the characteristics of signal transmission and processing in a neuron. Then, we discuss how and why enforcing neuronal diversity in a network is instrumental in enhancing the efficiency of neural networks, improving interpretability, and alleviating catastrophic forgetting. Next, we exhibit exemplary applications to show the benefits of introducing neuronal diversity in artificial neural networks. Lastly, we prospect future directions of neuronal diversity in artificial neural networks to further enrich the practice of NeuroAI.

\begin{figure}
\centering
\includegraphics[width=\linewidth]{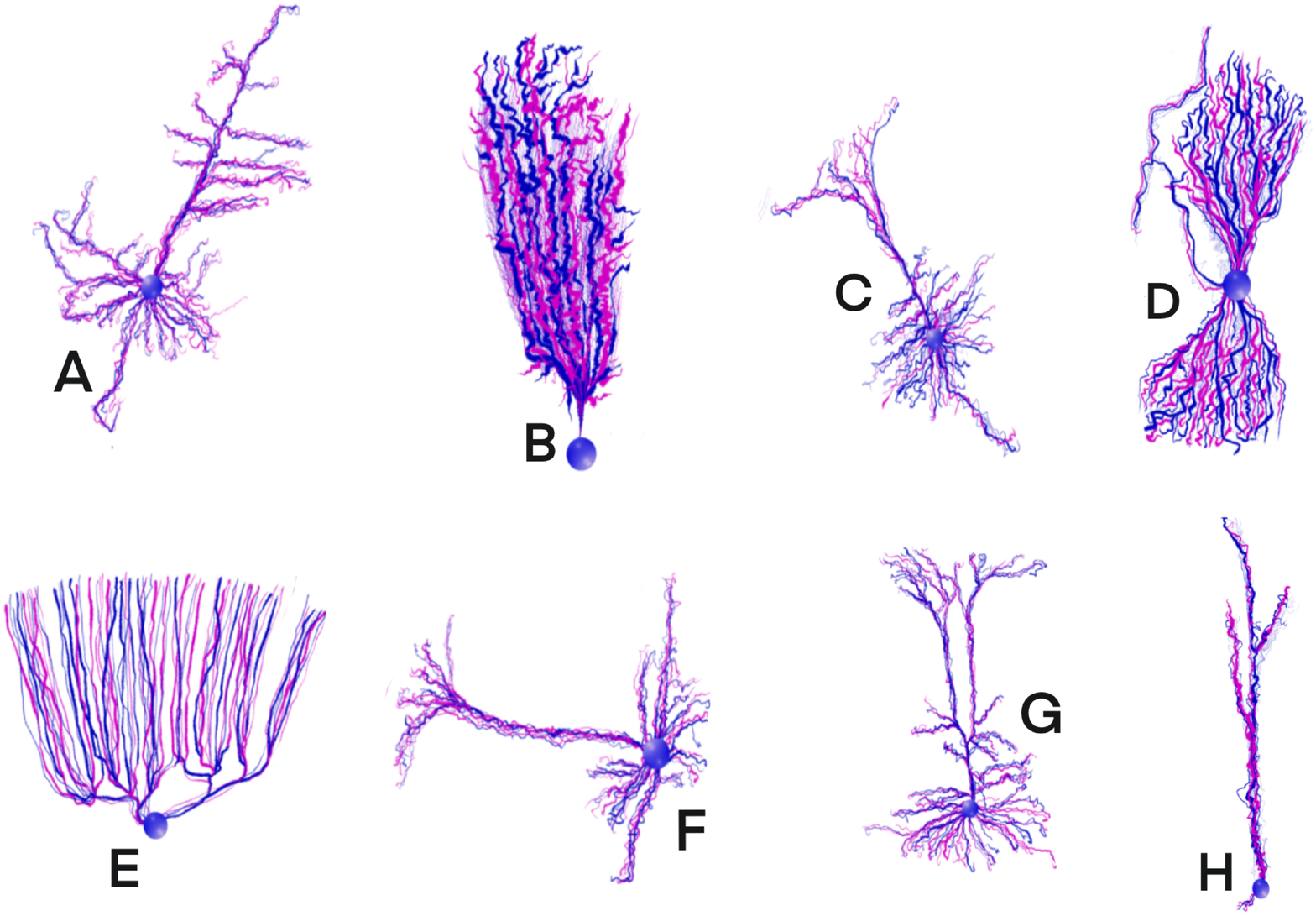}
\caption{Neurons exhibiting an extraordinary morphological diversity. The shape classifications of neurons include unipolar, bipolar, pseudounipolar, multipolar, and so on.  }
\label{fig:neuronal_diversity}
\end{figure}

\section{Biological Neuronal Diversity}

The extraordinary neuronal diversity originates from neuron differentiation \cite{flames2009gene}, a complicated process to obtain different types of neurons that motivates many molecular signals to drive electrophysiological, morphological, and transcriptional changes in a neuron. Neuronal diversity is reflected by different molecular, morphological, physiological, connectional features \cite{peng2021morphological}, and so on. Here, we briefly introduce morphological and functional diversity: the former is the most obvious diversity, and the latter is instrumental to understanding the different functionalities of a brain.

\textbf{Morphological diversity:} As a defining characteristic of neuronal types, morphological diversity, as shown in Figure \ref{fig:neuronal_diversity}, includes diversity in projection patterns, direction patterns, the density of branches, etc. According to the structural differences found in microscopy, neuroscientists previously have roughly divided neurons into four categories: unipolar, bipolar, multipolar, and pseudo-polar. Unipolar and pseudo-polar neurons have only one neurite extending from the soma, while that of a pseudo-polar neuron will soon split into two branches with a T-shaped structure directing to peripheral receptors and central spinal cords, respectively. While bipolar neurons extend protrusions from each end of the soma and evolve into dendrites and axons, respectively, multipolar neurons have one axon and several dendrites, which are the most common neuron type, constituting many complex central neural networks. Furthermore, more precise morphology studies reveal more concrete diversity from morphometry features to projection patterns on the regional level \cite{wang2021crhmorph,peng2021morphological}. 

\textbf{Functional diversity:} Functional diversity can be seen at structural, neuronal, network, system, and behavioral levels. Neuronal functional diversity is a natural result of evolution in order to conduct complex behaviors in human life. Let us take three basic functions as examples to explain.

$\bullet$ A human has five basic senses: sight, smell, touch, taste, and hearing, along with other senses including balance, proprioception, interior space emotion, time spiciness, etc. Hence, there exist sensory receptors that can specifically translate stimuli of different forms such as temperature, light, force, and sound, and then transmit impulses to the central nervous system. Once a sensory stimulus is applied to the exposed body part where the dendrite of the corresponding receptor neuron exists, the signal cascade and active signal pathway will be transferred, respectively. Then the functional sensory neural network starts to work. For example, as the human nose smells the flower, the olfactory sensory neurons located in the olfactory mucosa will be fired and evoke the olfactory network. Then a series of reactive events will happen.

$\bullet$ Brain regions are often tied with functionalities. There are multiple brain regions whose neurons are involved in motion. Those motor neurons usually correspond to certain muscles. The more agile the body parts are, the more delicate the movement is, and the bigger volume will the corresponding neural groups grow to. Despite the target diversity, neurons from different motor regions also differ in function with regard to the process and degree of motion. For example, neurons in the primary motor area will lead to the simple movement of body muscles on the opposite side after the stimulus. The premotor area related to the gross shrinkage of muscle is all over the body, responsible for some aspects of motor control such as movement preparation, sensory instruction, or direct control of certain movements. The supplementary motor area serves a number of motor suggestion functions including internally generated motor planning, sequence planning of movements, and coordination of both sides of the body. More subtle mechanisms of delicate motor have been studied in a long term with technique on different level \cite{pinto2021diversefunction}. 

$\bullet$ The GPS system in our brain is attributed to place neurons, grid neurons, direction neurons, boundary neurons, speed neurons, etc. These neurons coordinate together so that our brain knows where we are, where we are heading, and how far we move. When entering a particular place, a place neuron \cite{o1998place} is fired. However, place neurons alone cannot explain how a human navigates the environment. A coordinate system is established after grid neurons \cite{fyhn2008grid} are respectively activated, as one traverses a set of small regions. These roughly equal regions are arranged in a periodic array to cover the entire open environment. Together with boundary neurons \cite{lever2009boundarycells} encoding the space borderline, a mental map of the physical world is built, which allows for global precise positioning. In addition, neurons such as direction neurons \cite{taube1990head}, speed neurons \cite{gois2018characterizing}, and angular head velocity neurons \cite{stackman1998firing} inform the brain of characters of the motion in the environment and help modulate it. For example, when one makes a turn at a corner of a street, the direction neuron is fired to monitor the direction of movement. If one is in a hurry to arrive at the destination, the speed neuron is activated to supply the speed information.

The functional diversity of neurons is everywhere in the brain. In fact, a basic observation regarding the brain is that all complicated intelligence behaviors are a consequence of motivating different types of neurons. The above-mentioned different navigation neurons collaborate complementarily to realize all necessary functionalities for navigation.

\section{Characteristics of Biological Neurons}

Per the premise of NeuroAI, one should retrospect how the information is transmitted and processed throughout a biological neuron \cite{gardner2022cores} before introducing new neurons into artificial networks. To endow a network with more human-like capabilities, the neural mechanism should first be discussed. Roughly, it is divided into three steps: signal transduction (inbound), compartmentalized dendritic computation, and signal transduction (outbound). Our retrospection is centered around the macroscopic mechanisms and hallmarks relevant to neural computation. Unless necessary, we will not go into the level of cellular molecular biochemistry.

\textbf{Signal Transduction (Inbound):} 
Neurons can transduce almost all types of physical signals into electrical signals, \textit{e.g.}, optical \cite{bi2006ectopic,deisseroth2011optogenetics}, mechanical \cite{suchyna2007mechanosensitive}, biochemical \cite{rosenbaum2009structure}. Within the biological neural network, the most common is biochemical signal transduction.
This transduction of a neuron is carried out by two categories of transducers of the current neuron: Membrane receptor-mediated type and nuclear receptor-mediated type. Two classic subcategories of the former are ligand-gated ion channels and transmembrane G-protein coupling receptors. The former \cite{hucho2001ligand} will open to allow the ion flux via a conformational change when the receptor binds a chemical messenger such as neurotransmitters; the latter \cite{gilman1987g} will bind with neurotransmitters and activate coupling G proteins on the membrane, which either directly causes the ion channel to open via protein-protein interaction or activates the enzyme that expedites the synthesis of the intracellular second messenger towards a lagged \cite{lohse2008kinetics}, cascaded \cite{lamb1992g}, and longer-lasting \cite{sheng1990membrane} modulation\footnote{https://openbooks.lib.msu.edu/neuroscience/chapter/neurotransmitter-action-g-protein-coupled-receptors/} to ion channels. Finally, an electrical signal will be stimulated and further processed.

\textbf{Dendritic Computation:} In the early days, dendritic trees are believed to solely receive and transport information to the axon (passive). The pioneer of modern neuroscience once asked "Why do dendritic trees even exist"\cite{branco2010single}? However, with the progress of sharp electrodes \cite{ling2007history}, ever-growing evidence suggests that the dendritic branches are compartmentalized functional units (active). They play a fundamental role in many key computations such as coincidence detection, detection of motion direction, and storage of multiple input features (Chapter 15, \cite{stuart2016dendrites}). Thus, a dendritic branch is a compartmentalized computation unit enabled and defined by the passive and active properties. The integrated dendritic computation, therefore, is more complicated and needs better understanding.

\textit{Passive Properties}: Due to the intrinsic resistance, the most salient passive property of dendrites is its attenuation effect whose role is to keep the signal local and sparse. The attenuation rate decreases with the diameter and increases with the length \cite{rall1959branching,rall1960membrane}. A branchpoint, which is a bifurcation between the apical trunk and oblique dendrite, also serves as a strong attenuator to restrict the active signal transmission. In addition to the attenuation effect, dendritic morphology can affect the fire rate and fire pattern as well \cite{mainen1996influence,ferrante2013functional}. Furthermore, the signal propagates into distal dendritic arbors more readily than towards the soma (Chapter 14, \cite{stuart2016dendrites}). Such an asymmetry can empower the neuron with the ability to infer the direction of motion based on the activation order of the somatic response. 
Finally, the dendritic structure is shown to facilitate the sparse coding in biological networks \cite{de2015dendritic} that are associated with energy efficiency \cite{wang2020relationship}, feature discrimination \cite{chavlis2017dendrites}, and memory capacity \cite{brunel2004optimal}.

\textit{Active Properties}: A dendrite is embedded with a variety of ionic channels that enable powerful signal processing abilities and assist the information transmission intracellularly. 

$\bullet$ Voltage-gated channels are nonlinear functional devices. For example, no action potential is generated when the input current is below a certain threshold. When the input current exceeds the action potential threshold, plenty of ion channels will open and generate an abrupt change in membrane potential. The action potential cancels the attenuation effect of the morphology and ensures the reliable propagation of the signal \cite{levitan2002neuron}. The dendritic spikes can exhibit other complicated input-output relations other than the all-or-none relation. Recently, Gidon et al. \cite{gidon2020dendritic} showed that the human layer 2/3 cortical neurons produce maximal activation when a stimulus is close to zero, and the activation is dampened when the stimulus goes stronger. Interaction between ion channels adds another layer of nonlinearity, \textit{e.g.}, calcium-dependent potassium channels \cite{vergara1998calcium} link potassium channels with calcium channels, which realizes a more precise regulation for neuronal excitability. 

$\bullet$ Voltage-gated channels are highly discriminatory, which enhances the dendrites' computation power. Typically, voltage-gated channels are only open to one kind of ions over another. As such, those channels are named after the most easily passed ions. Furthermore, for the same type of ion channels, there exist a variety of subtypes that differ in their voltage dependence, kinetics, single-channel conductance, and so on. Different ion channels present different firing patterns \cite{levitan2002neuron}.

$\bullet$ Active dendrites respond in a location-dependent and time-dependent manner, implicating its spatiotemporal processing ability \cite{cook1997active,das2014active}. For example,
dendritic sodium spikes are propagatable throughout the dendritic tree \cite{ariav2003submillisecond}, while the distal apical trunk tends to initialize calcium spikes that can only spread to the apical dendritic tree \cite{schiller1997calcium,polsky2004computational}. In addition, both sodium and potassium spikes participate in coincidence detection \cite{ariav2003submillisecond,losonczy2006integrative,larkum1999new}. For example, the potassium spike was activated when distal and proximal dendritic regions of cortical neurons are synchronously activated \cite{larkum1999new}.

\begin{figure}
\centering
\includegraphics[width=\linewidth]{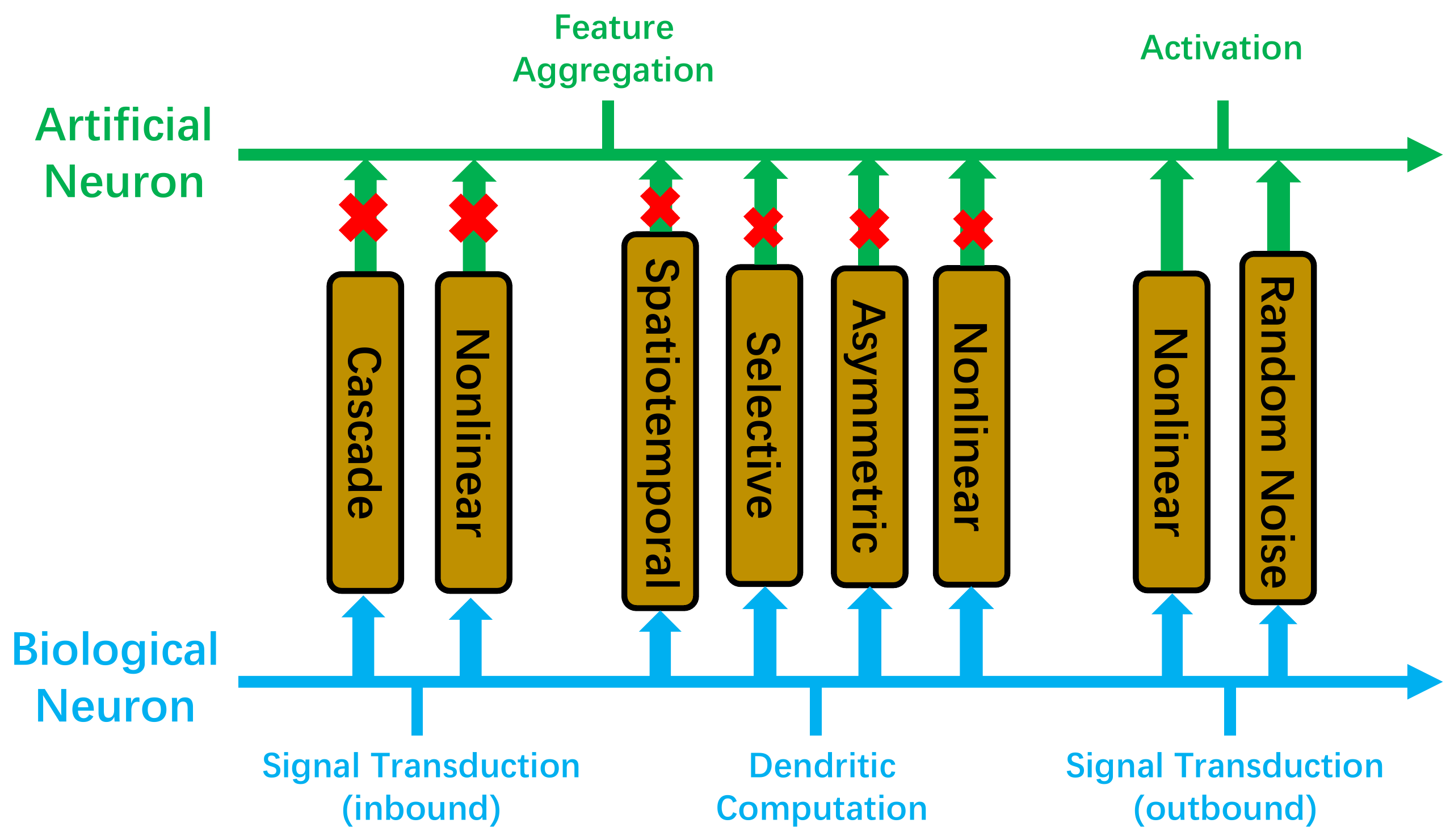}
\caption{A schematic plot to show what kinds of computational properties of biological neurons remain unseen in the current mainstream artificial neurons.}
\label{fig:summary}
\vspace{-0.4cm}
\end{figure}

\begin{table*}
\caption{A summary of different types of activation functions. }
\centering
\scalebox{1}{
\begin{tabular}{l|l|c}
\hline
    \hline
    \rowcolor{LightCyan}
    \textbf{Type} & \textbf{Representative Example} & \textbf{Mathematical Expression} \\ 
    \hline
    \rowcolor{Gray}
    \multirow{1}{*}{Sigmoid/Tanh Type}  &  Sigmoid \cite{han1995influence}  & $\frac{1}{1+e^{-x}}$  \\  
    \rowcolor{Gray}
    & PSF \cite{chandra2004activation} & $\frac{1}{(1+e^{-x})^m}$ \\
    \rowcolor{Gray}
    & ISigmoid \cite{qin2018optimized} & $ISigmoid(x) = \left\{\begin{array}{cc} a(x-a)+1/(1+e^{-a}),&   x \leq -a\\ 1/(1+e^{-x}), & -a < x < a \\ a(x+a)+1/(1+e^{-a}), & x \geq a \end{array}\right.$ \\
    \rowcolor{Gray}
& Tanh \cite{maas2013rectifier} & $(e^x - e^{-x})/(e^x + e^{-x}) $ \\
\rowcolor{Gray}
& scaled Tanh \cite{lecun1998gradient} & $A (e^{Bx} - e^{-Bx})/(e^{Bx} + e^{-Bx})$ \\
     \hline
     \rowcolor{Gray2}
    \multirow{1}{*}{ReLU Type}  &  ReLU \cite{nair2010rectified}  & $ReLU(x)=\left\{\begin{array}{cc} x,&   x \leq 0\\ 0, & x<0 \end{array}\right.$  \\  
    \rowcolor{Gray2}
    & Leaky-ReLU \cite{maas2013rectifier} & $LeakyReLU(x)=\left\{\begin{array}{cc} x,&   x \geq 0\\ \alpha \cdot x, & x<0 \end{array}\right.$ \\
    \rowcolor{Gray2}
    & Concatenate-ReLU \cite{shang2016understanding}  & $[ReLU(x), ReLU(-x)]$\\
    \rowcolor{Gray2}
     & GenLU \cite{fan2020soft}  & $GenLU(x) = sgn(x) \max\{|x| + b, 0\}$\\
     \rowcolor{Gray2}
     & Randomly Translational ReLU \cite{cao2018randomly} & $RTReLU(x)=\left\{\begin{array}{cc} x+a,&   x+a \geq 0\\ 0, & x+a\leq 0 \end{array}\right.$ \\
     \rowcolor{Gray2}
     &  ELU \cite{clevert2015fast}  & $ELU(x)=\left\{\begin{array}{cc} x,&   x \geq 0\\ \alpha(e^x-1), & x<0 \end{array}\right.$  \\ 
     \rowcolor{Gray2}
     & SReLU \cite{jin2016deep} & $SReLU(x)= \left\{\begin{array}{cc} t^r + a^r(x-t^r),&   x \geq t^r\\  x, & t^l \leq x \leq t^r \\ 
     t^l + a^l (x-t^l) , & x \leq t^l \end{array}\right.$ \\
     \rowcolor{Gray2}
     & Natural Logarithm ReLU \cite{liu2019natural} & $NTReLU(x) =\ln(\beta\max\{0, x\} + 1)$\\
     \hline  
       \rowcolor{Gray3}    
    \multirow{1}{*}{Radial Basis Function Activation}  &  Gaussian \cite{moody1989fast}  &  $\rho(\x) = \exp(\beta \Vert \x-c \Vert^2)$\\ 
    \hline
     
      \rowcolor{Gray4}    
    \multirow{1}{*}{Bioplausible Activation}  &  apical dendrite activation (ADA) \cite{georgescu2020non} &  $ADA(x)=l\min \{0, x\}+\max\{0, x\}\exp(-\alpha x+ c)$\\  
    \rowcolor{Gray4}  
    & Bionodal root unit (BRU) \cite{bhumbra2018deep} &  $BRU(x)=\left\{\begin{array}{cc} (r^2 x +1)^{\frac{1}{r}},&   x \geq 0\\ e^{rz}-\frac{1}{r}, & x<0 \end{array}\right.$ \\  
    \hline  

      \rowcolor{Gray5}    
    \multirow{1}{*}{Noisy Activation}  &  \cite{gulcehre2016noisy} &  $\phi(x)=h(x)+s(x)$,  where $s(x)$ is random noise.   \\
    \hline  
    
\end{tabular}}
\label{tab:statistics}
\end{table*}

\textbf{Signal Transduction (Outbound):}
Transmitting information from a neuron's interior to its exterior has two fundamentally different modes. One is the direct electrical transfer via a specialized interconnected connection called the gap junction \cite{faber2018two}. The transfer through gap junctions is very fast. The other mode is chemical transfer via the synapses, \textit{i.e.}, the presynaptic neuron releases the so-called neurohormone \cite{nelson2005introduction} or neurotransmitter \cite{lodish2000neurotransmitters} that will diffuse to the target neuron. The chemical transfer converts the electrical signal into the chemical release.

$\bullet$ The chemical transfer can exert different effects, depending on the types of released neurotransmitters \cite{eccles2013physiology}: The excitatory neurotransmitters fire the target neuron, while the inhibitory ones inhibit the target neuron, and modulatory neurotransmitters affect the effects of other chemical messengers. In most situations, one synapse can only be either excitatory or inhibitory. 

$\bullet$ The chemical transfer is subjected to random noise \cite{levitan2002neuron}. Even when the action potential is absent, a small depolarization is recorded in the target neuron due to the spontaneous random release of a small number of neurotransmitters from the presynaptic terminal. 

$\bullet$ The relation between the level of action potentials and the amount of the released neurotransmitter is nonlinear. When the voltage is low, only a few calcium channels are open. The high calcium concentration is achieved near those open channels, thereby only facilitating the fusion of nearby vesicles. When the voltage slightly increases, more channels open; therefore, a relatively small and uniform calcium concentration emerges. But the calcium level may not be sufficiently high to trigger subsequent processes \cite{poage2002presynaptic}.

Overall, despite the complicated information processing of a biological neuron, several salient characteristics should be noted: i) nonlinearity ubiquitously exists in every stage of information processing, not just in membrane channels; ii) a dendrite not only serves information transmission but also serve as compartmentalized computation unit; iii) neurons have the spatiotemporal information ability.

Figure \ref{fig:summary} shows what kinds of computational properties of biological neurons remain unseen in the current mainstream artificial neurons. We divide the operations of artificial neurons into two stages: feature aggregation and nonlinear activation. It is widely believed that the nonlinear activation function in artificial neurons corresponds to the voltage/ligand-gated channels of biological neurons. However, in biological neurons, except for the outbound signal transduction, the roles of voltage/ligand-gated channels are to assist the information transmission and processing intra-cellularly, instead of emitting information extra-cellularly. Therefore, we think that the outbound signal transduction of biological neurons corresponds to the nonlinear activation of artificial neurons, while the inbound signal transduction and dendritic computation together are in analogy to the feature aggregation of artificial neurons.


\section{Neuronal Diversity in Artificial Neural Networks} 

In this section, we classify the studies of introducing neuronal diversity into artificial networks into four categories: activation design, polynomial neurons, dendritic neurons, and spiking neurons. Note that complex-valued neurons \cite{bassey2021survey,zhang2022towards} are not included because although the complex-valued neuron is novel, its idea is based on the need of addressing complex-valued data, thereby its basic computation (inner product + nonlinear activation) is the same as the real-valued neuron. Now let us introduce four categories in detail:

\textbf{Activation Design.} Nowadays, the mainstream neuron type is to compute the inner product between the input and the connectivity parameters of upper neurons followed by a nonlinear activation function, which is mathematically formulated as 
\begin{equation}
    \sigma(\x^\top\mathbf{w}+b),
\end{equation}
where $\x$ is the input, $\mathbf{w}, b$ are parameters, and $\sigma$ is the activation function of current neuron. Past years have witnessed a plethora of novel activation functions being designed \cite{dubey2021comprehensive}. Three different properties are often considered in developing a novel activation function: i) As mentioned earlier, an activation function should be nonlinear, which is a necessity to guarantee a network has a sufficient representation and discrimination ability. Nonlinearity is also bioplausible, \textit{i.e.}, the outbound signal transduction in a biological neuron is nonlinear. ii) An activation function should allow a normal gradient flow across layers when a network is deep, \textit{i.e.}, no gradient vanishment and explosion. iii) It should facilitate information transmission to expedite the extraction of useful features from data.

$\bullet$ Logistic Sigmoid/Tanh: The logistic sigmoid \cite{han1995influence} was extensively used in the early stage of neural networks. However, the employment of such an activation function in a deep network suffers gradient vanishment and poor convergence. The gradient is killed when the pre-activation value is super high or low, while the non-zero-centered nature forces the convergence trajectory to go zig-zag \cite{dubey2021comprehensive}. Tanh is of zero-centric nature, but its computational complexity is high, and still subjected to the gradient issue. Several variants of Tanh were proposed to enlarge the range of output function \cite{lecun1998gradient} and overcome the gradient vanishment problem \cite{chandra2004activation,gulcehre2016noisy,qin2018optimized}.

$\bullet$ Rectified Linear Unit (ReLU): ReLU \cite{nair2010rectified} is currently the most popular activation function thanks to its excellent scalability.
However, some researchers argued that the employment of ReLU causes the loss of useful information in negative parts. To address this issue, several variants that allow the passage of negative parts were proposed such as Leaky-ReLU \cite{maas2013rectifier}, Concatenate-ReLU \cite{shang2016understanding}, GenLU \cite{fan2020soft}, Randomly Translational ReLU \cite{cao2018randomly}, ELU \cite{clevert2015fast}. Besides, ReLU was believed to have a limited discriminative ability because it just assumes a linear relation in the positive range. Therefore, several variants were proposed to modify ReLU towards an enhanced non-linearity such as SReLU \cite{jin2016deep} and Natural Logarithm ReLU \cite{liu2019natural}. 

$\bullet$ Radial Basis Function (RBF): A radial basis function activation is usually used in the radial basis function network \cite{moody1989fast}, which is formulated as 
\begin{equation}
    \psi(\x)=\sum_{i=1}^N a_i \rho(\Vert \x-c_i \Vert),
\end{equation}
where $\rho$ is often taken as a Gaussian function $\rho(\Vert \x-c_i \Vert) = \exp(-\beta_i \Vert \x-c_i \Vert^2)$. The radial basis function network can be regarded as a fuzzy logic system such as the Takagi-Sugeno rule system \cite{takagi1985fuzzy} whose rule is of the format: “if $x \in \mathrm{set}~A$ and $y \in \mathrm{set}~B$, then $z=f(x,y)$” \cite{park1991universal}.

$\bullet$ Bioplausible Activation: One notable class of activation functions are those biologically-inspired activation functions. Georgescu \textit{et al.} \cite{georgescu2020non} proposed a bioplausible activation function by mathematically modeling the newly-discovered action potential pattern in a study \cite{gidon2020dendritic} published in Science. Bhumbra \cite{bhumbra2018deep} introduced a bionodal root activation based on the input-output relation acquired from physiological measurement. Electrophysiological recordings show that only a moderate increment in inputs is required to drive an action potential for some neurons \cite{bhumbra2018deep}. Then, after an initial linear relation, the input-output curve partially saturates because the voltage-gated channels become less sensitive.

$\bullet$ Noisy Activation: The idea of noisy activation is to inject noise into the activation function when it is saturated. Such an injection can make gradients flow easily \cite{gulcehre2016noisy}.  

\begin{figure*}
\centering
\includegraphics[width=\linewidth]{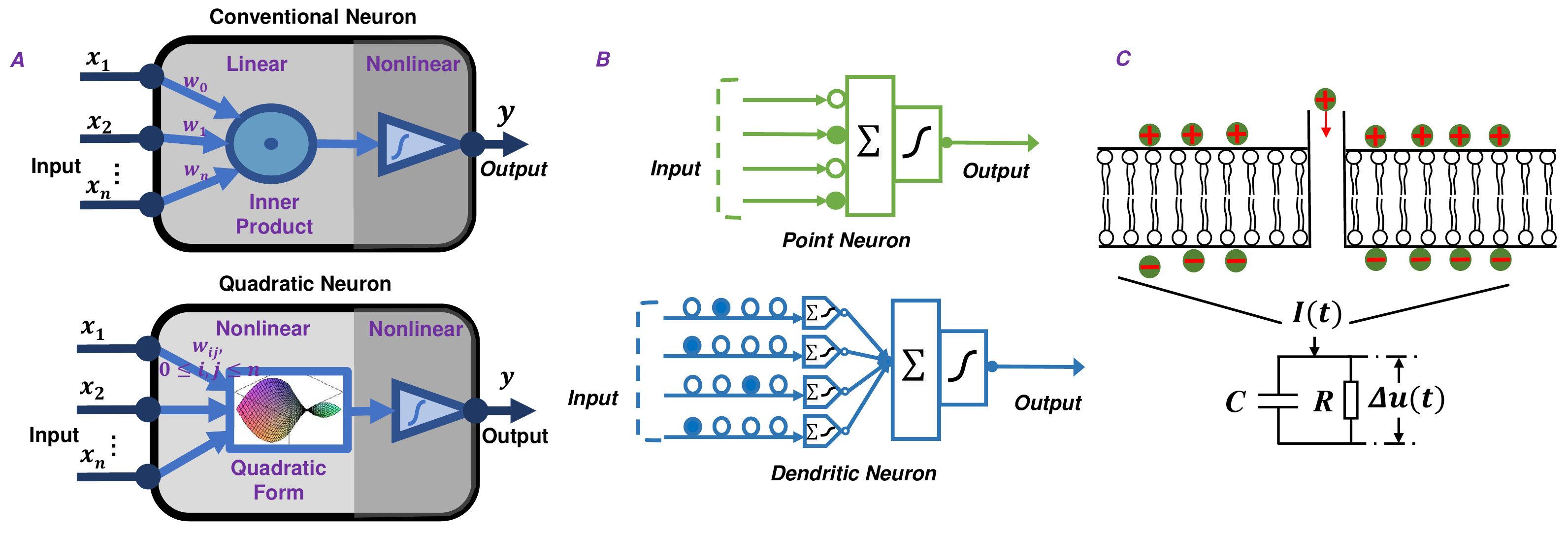}
\caption{Three exemplary new types of neuron designs: polynomial neurons, dendritic neurons, and spiking neurons. Polynomial neurons, dendritic neurons, and spiking neurons conform to the three salient characteristics of biological neurons i) ubiquitous nonlinearity in information processing; ii) the active dendrite; iii) the spatiotemporal information ability, respectively.} 
\label{fig:three_in_one}
\end{figure*}

\begin{table}
\centering
\caption{An overview of recent polynomial neurons. $\sigma(\cdot)$ denotes the nonlinear activation function. $\odot$ denotes Hadamard product. Here, we omit the bias terms in a neuron, and $\W, \w_i$ denotes learnable weight matrix and vectors, respectively.}
\begin{tabular}{|l|c|}
\hline
Work Reference           & Formula          \\ \hline
\makecell[l]{Zoumpourlis \textit{et al.}(2017) \cite{zoumpourlis2017non} \\ Cheung\&Leung (1991) \cite{cheung1991rotational}}  & $\sigma(\x^{\top}\W\x+\x^\top \w_1)$ \\
\hline
Tsapanos \textit{et al.} (2018) \cite{tsapanos2018neurons} & $\sigma\Big((\x^\top\mathbf{w}_{1}+b_{1})(\x^\top\mathbf{w}_{2}+b_{2})\Big)$ \\
\hline
Fan \textit{et al.}(2018)   \cite{fan2018new}      & \makecell{$\sigma\Big((\x^\top\mathbf{w}_{1}+b_{1})(\x^\top\mathbf{w}_{2}+b_{2})+$ \\$(\x\odot\x)^\top\mathbf{w}_{3}\Big)$ }       \\ \hline
\makecell[l]{Redlapalli et al. (2003)\cite{redlapalli2003development}\\
Jiang \textit{et al.} (2019) \cite{jiang2020nonlinear}\\ Mantini\&Shah (2021) \cite{mantini2021cqnn}}     & {$\sigma(\x^{\top}\W\x$)} 
                       \\ \hline
Goyal \textit{et al.}(2020) \cite{goyal2020improved}    & $\sigma\Big((\x\odot\x)^\top\mathbf{w}^{b}+c\Big)$              \\ \hline
       \makecell[l]{Xu \textit{et al.}(2022) \cite{xu2022quadralib} \\Bu\&Karpatne(2021) \cite{bu2021quadratic}}  & {$\sigma((\x^\top\w_1) (\x^\top\w_2)+\x^\top\w_3)$}                             \\ \hline
\end{tabular}
\label{tab:neurons}
\end{table}

\textbf{Polynomial Neuron.} Compared to the activation function, altering feature aggregation is much less explored. Per our earlier analysis, the nonlinearity in biological neurons is embedded not only in the outbound signal transduction but also in the inbound signal transduction and dendritic computation. But the inner product is a linear operation, which does not align with nonlinearity of the inbound signal transduction and dendritic computation. The use of polynomial neurons is to endow an artificial neuron with the nonlinear feature processing ability in the aggregation phase to better fit this strongly nonlinear information world. The story of polynomial neurons originates from the Group Method of Data Handling (GMDH \cite{ivakhnenko1971polynomial}), which takes a high-order polynomial as a feature extractor:
\begin{equation}
\begin{aligned}
      &   Y(\x_1,\cdots,\x_n) \\
     = & \sum_i^n a_i \x_i + \sum_i^n \sum_j^n a_{ij} \x_i \x_j  + \sum_i^n \sum_j^n \sum_k^n a_{ijk} \x_i \x_j \x_k + \cdots ,
\end{aligned}
\end{equation}
where $\x_i$ is the $i$-th input, and $a_i, a_{ij}, a_{ijk},\cdots$ are coefficients for interaction terms. Usually, only quadratic terms are retained in this model to avoid nonlinear explosion for high-dimensional inputs. Furthermore, with GMDH, the so-called higher-order unit was defined in \cite{poggio1975optimal,giles1987learning,lippmann1989pattern} which is mathematically formulated as
\begin{equation}
    y=\sigma(Y(\x_1,\cdots,\x_n)), 
\end{equation}
where $\sigma(\cdot)$ is a nonlinear activation function. To achieve a balance between maintaining the power of high-order units and parameter efficiency,  Milenkoiv \textit{et al.} \cite{milenkovic1996annealing} only utilized linear and quadratic terms and proposed to use an annealing technique to find optimal parameters.

Recently, high-order, particularly quadratic units were revisited \cite{zoumpourlis2017non,tsapanos2018neurons,chrysos2021deep,livni2014computational,krotov2018dense,xu2022quadralib}. In the work by Chrysos \textit{et al.} \cite{chrysos2021deep}, the complexity of higher-order units as described by Eq. \eqref{HOunits} were greatly reduced via tensor decomposition and factor sharing; therefore, they scaled polynomial networks into a very deep paradigm to achieve the cutting-edge performance on several tasks. In \cite{zoumpourlis2017non,jiang2020nonlinear,mantini2021cqnn}, a quadratic convolutional filter of the complexity $\mathcal{O}(n^2)$ was proposed to replace the linear filter. In \cite{tsapanos2018neurons}, a parabolic neuron: $\sigma\Big((\x^\top\mathbf{w}_{1}+b_{1})(\x^\top\mathbf{w}_{2}+b_{2})\Big)$ was proposed for deep learning, while in \cite{goyal2020improved}, $\sigma\Big((\boldsymbol{x}\odot\boldsymbol{x})^\top\mathbf{w}\Big)$ was proposed.
Fan \textit{et al.} \cite{fan2018new} proposed a simplified quadratic neuron with $\mathcal{O}(3n)$ parameters: $\sigma\Big((\x^\top\mathbf{w}_{1}+b_{1})(\x^\top\mathbf{w}_{2}+b_{2})+(\x\circ\x)^\top\mathbf{w}_{3}+c\Big)$ and further argued that higher-order neurons are not necessary because the fundamental theorem of algebra suggests that any polynomial can be factorized into a product of linear and quadratic terms \cite{remmert1991fundamental}. Neuron designs in \cite{tsapanos2018neurons,goyal2020improved} are special cases of that in \cite{fan2018new}.
Bu \textit{et al.} \cite{bu2021quadratic} utilized the quadratic neuron $\sigma\Big((\x^\top\mathbf{w}_1)(\x^\top\mathbf{w}_2)+\x^\top\mathbf{w}_3\Big)$, which is equivalent to \cite{tsapanos2018neurons} when combining $\x^\top\mathbf{w}_3$ into $(\x^\top\mathbf{w}_1)(\x^\top\mathbf{w}_2)$. Xu \textit{et al.}'s quadratic neuron design \cite{xu2022quadralib} is the same as \cite{bu2021quadratic}. Liu and Wang \cite{liu2021dendrite} defined the so-called Gang neuron that is recursively formulated as $\mathbf{A}^{l} = \mathbf{W}^{l,l-1}\mathbf{A}^{l-1}\circ \x$ and $A^0 = \x$, which is essentially a polynomial neuron under a particular tensor decomposition.

\textbf{Dendritic Neuron.} The current mainstream neuron type is a point neuron which computes a single weighted sum of all synapses. Such modeling actually ignores the active computation of the dendritic fibers and only takes them as a passive transmission medium. However, as aforementioned, dendrites are active and serve as compartmentalized information processing units to assist a neuron to perform different kinds of tasks. Thus, \textit{can we go from a linear and passive neuron (point neuron) to a nonlinear and active neuron with extensive dendrite branching and extra nonlinear computation?} (Chapter 14, \cite{stuart2016dendrites}). Along this direction, a feasible way to design a dendritic neuron is to first compute a subset of nonlinear terms within a subset of computational units, then combine the responses of these units and nonlinearly map their total response (Chapter 16, \cite{stuart2016dendrites}), as shown in Figure \ref{fig:three_in_one}. For example, dendritic branches were simulated to compute the sum of products (pi-sigma units \cite{mel1987murphy,durbin1989product}) and to implement Boolean logic networks \cite{koch1982retinal,shepherd1987logic,zador1991nonlinear}. Shin \textit{et al.} reported the so-called pi-sigma unit \cite{shin1991pi},
\begin{equation}
    h_{ji} = \sum_k \omega_{kji} \x_k + \theta_{ji} \ \ and \ \ y_i = \sigma(\prod_j h_{ji}), 
\label{HOunits}    
\end{equation}
where $h_{ji}$ is the output of the $j$-th sigma unit for the $i$-th output element $y_i$, and $\omega_{kji}$ is the weight of the $j$-th sigma unit associated with the input element $\x_k$. The cluster-sensitive phenomenon was discovered that given a fixed number of synaptic inputs, concentrating the activated synapses of intermediate size can lead to the largest post-synaptic response \cite{mel1991clusteron,gasparini2004initiation,polsky2009encoding}. Based on this phenomenon, the "clusteron" was proposed in \cite{mel1991clusteron} whose output is given by
\begin{equation}
    y = g(\sum_{i=1}^N a_i),
\end{equation}
where $a_i=w_i x_i (\sum_{j\in D_i} w_j x_j)$ is the net excitatory input at synapse $i$, and $D_i=\{i-r,i-r+1,\cdots,i,\cdots,i+r-1,i+r\}$ is a set of neighbors of the synapse $i$. It can be seen that the clusteron is a constrained sigma-pi unit with products of neighboring synapses. Furthermore, Jadi et al. \cite{jadi2014augmented} proposed a two-stage dendritic neuron model:
\begin{equation}
    y=\mathrm{FI}(\sum_j W_j d_j),
\end{equation}
where $\mathrm{FI}$ is an experimentally-determined frequency-current relation, and $d_j=\sigma(\sum_{i} w_{ij}x_i)$. However, such a dendritic neuron is essentially isomorphic to a two-layer network. Hawkins and Ahmad \cite{hawkins2016neurons} and Grewal \textit{et al.} \cite{grewal2021going} developed a type of dendritic neurons, as shown in Figure \ref{fig:three_in_one}, where each dendritic branch contains multiple synapses such that each branch can detect multiple input patterns. At the same time, Grewal \textit{et al.} showed that such a type of dendritic neurons can alleviate catastrophic forgetting. In the aspect of hardware, Li \textit{et al.} \cite{li2020power} experimentally demonstrated that neural networks with artificial dendrites are power-efficient.

\textbf{Spiking Neuron.} The current mainstream neuron types are regarded as the second-generation neurons that can only process static amplitude information. The neurons that have the spatiotemporal information ability are referred to as the third-generation neurons, which are primarily of the "integrate-and-fire" type \cite{izhikevich2003simple,hodgkin1952quantitative,burkitt2006review} via spikes \cite{roy2019towards}.

\begin{figure*} 
\centering
\includegraphics[width=0.75\linewidth]{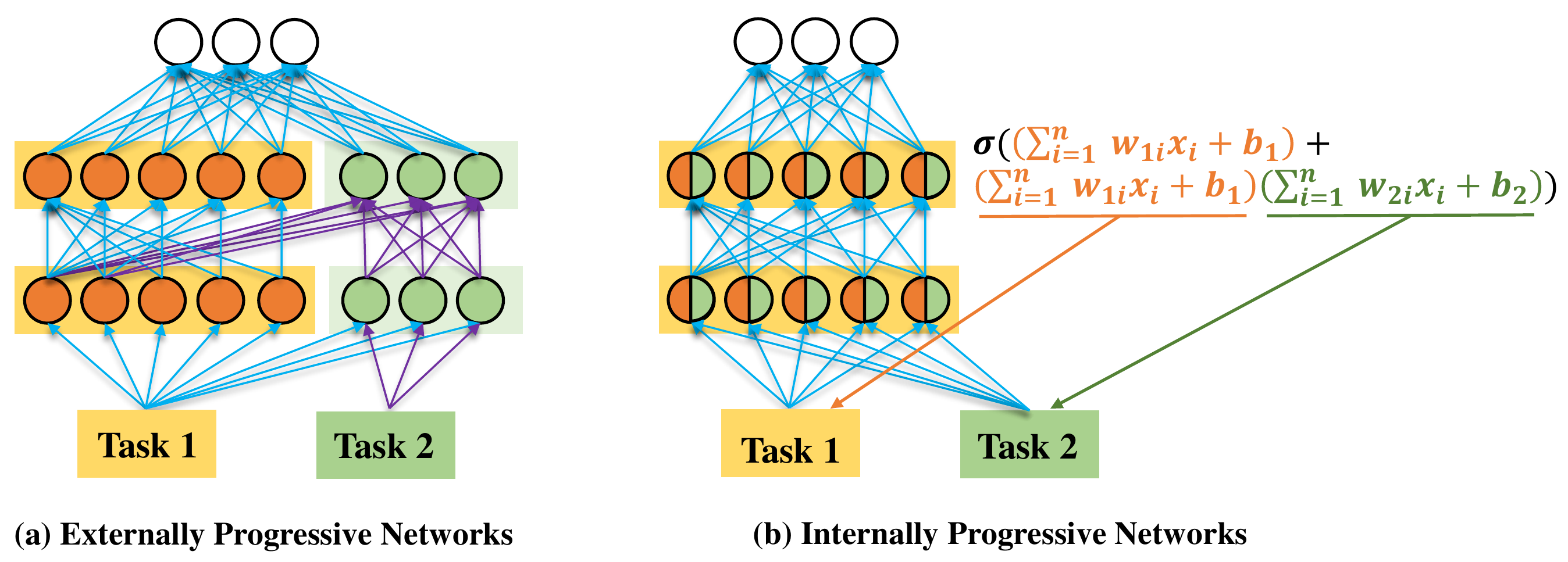}
\caption{Polynomial networks can work progressively to avoid catastrophic forgetting.} 
\label{fig:internally_progressive_network}
\vspace{-0.3cm}
\end{figure*}

For example, the leaky integrator neuron is shown in Figure \ref{fig:three_in_one}. Its neuronal dynamics consists of two parts: (i) an equation that describes the change of the transmemebrane potential difference; (ii) a mechanism to generate a spike. We use the law of current conservation to derive the integration equation of spiking neurons. The current is split into two components:
\begin{equation}
    I_t=I_R+I_C = \frac{\Delta u_t}{R} + C \frac{d}{d t} \Delta u_t,
    \label{eqn:integrate_and_fire}
\end{equation}
where $\Delta u_t$ is the transmembrane potential difference, $I_t$ is the current from synapses or the external injection independent of the membrane potential, $R$ is the leaky resistance, and $C$ is the membrane capacitance. 
When $\Delta u_t$ hits a threshold, it forms a spike and then set to zero. Assuming a constant external current injection and no synaptic current, we have the following:
\begin{equation}
    \Delta u_t = I R (1-e^{-\frac{t}{RC}}).
\end{equation}
Next, the mechanism of generating a spike is via a spike response model \cite{gerstner1995time}, which simulates the refractory properties of a neuron because the membrane potential depends on the time of the last spike. There are other variants of the integrate-and-fire model: the linear integrate-and-fire neuron \cite{fusi1999collective} which replaces $\frac{\Delta u_t}{R}$ in Eq.\eqref{eqn:integrate_and_fire} with a constant term; the quadratic integrate-and-fire neuron \cite{ermentrout1996type,brunel2003firing,clusella2022kuramoto} which adds a quadratic term $u_t^2$ to the right side of Eq. \eqref{eqn:integrate_and_fire}; the exponential integrate-and-fire neuron \cite{fourcaud2005dynamics} which adds an exponential term regarding $u_t$ to the right side of Eq. \eqref{eqn:integrate_and_fire}.
As mentioned earlier, the information transmission in a neuron is coupled with noise. For example, the voltage-gated channels randomly open and close, and the vesicles randomly fuse with a neuron's membrane to release neurotransmitters. Thus, a diffusion variant of spiking neurons \cite{tuckwell1988nonlinear,tuckwell1980accuracy} was established by incorporating the stochastic nature of the current:
\begin{equation}
    I_t = \mu + \sqrt{2RC}\xi_t = \frac{\Delta u_t}{R} + C \frac{d}{d t} \Delta u_t,
\end{equation}
where $\mu$ is the average synaptic current, and $\xi_t$ is a Gaussian noise.

The second-generation neurons are trained with gradient descent, whereas the training of spiking neurons is difficult due to the non-differentiability of the spiking neurons. Trainability has been a major bottleneck in the development of spiking networks. Now there are roughly three approaches to train a spiking network. 

$\bullet$ The idea of conversion-based methods is to map a well-performed conventional network into a spiking network, which requires parameter recalibration and activation rescaling \cite{cao2015spiking,sengupta2019going,rueckauer2017conversion}. This method helps a spiking network to showcase the state-of-the-art performance on the ImageNet. However, such a conversion may be unfaithful when time steps are low. 

$\bullet$ The second method is to employ surrogate gradients \cite{lee2016training,gu2019stca,li2021differentiable}. The spiking network is hard to train because spiking neurons fire discrete spikes that are non-differentiable. To solve this problem, differentiable activation functions are employed as a surrogate. Moreover, unrolling a spiking neuron in a discrete, recursive form perfectly corresponds to a recurrent neural network. As a result, the training of a spiking network can be done via backpropagation through time (BPTT) \cite{werbos1988generalization}. The caveat of surrogate gradient methods is that it is computationally expensive for a large time step.

$\bullet$ Spike-timing-dependent plasticity (STDP) approaches \cite{bi1998synaptic,iakymchuk2015simplified,lobov2020spatial} gain great interest recently. SDTP methods are with local learning, following the Hebbian learning rule and updating weights in an unsupervised manner. However, STDP methods are hard to scale to multilayer spiking networks, and have limited performance on large-scale datasets.

\textbf{Remarks.} Polynomial neurons, dendritic neurons, and spiking neurons conform to the three salient characteristics of biological neurons i) ubiquitous nonlinearity in information processing; ii) the active dendrite; iii) the spatiotemporal information ability, respectively. But polynomial neurons, dendritic neurons, and spiking neurons are not exactly the same to the ways of biological neurons. We argue that artificial networks should draw inspiration instead of copying from neuroscience. In other words, designing new neurons should prioritize the need for real-world problems instead of blindly imitating biological neurons.

\section{Potential Gains}

So far, we have discussed new types of neurons, with the motivation of filling the gap between the current mainstream neuron type and the real-world biological neuron. Next, more importantly, \textit{what can we gain from these new neurons? \textit{i.e.}, how can neuronal diversity truly bring benefits to the aforementioned critical issues in artificial networks?} In the following, let us illustrate the potential gains of incorporating neuronal diversity into artificial networks. 

\textbf{Efficiency:} In the information world, most learning tasks establish a nonlinear mapping. Intuitively, it is more efficient to incorporate a nonlinear computation unit to learn a nonlinear function. Although the existing mainstream neurons can do the universal approximation when connected to a network, it is computationally heavy to use mainstream neurons to represent other neuron types because such a representation needs far more neurons, compared to directly adopting other neuron types in building a model. It was proved that there exists a class of functions that can be approximated by a heterogeneous network made of both quadratic and conventional neurons with a polynomial number of neurons, but is hard to approximate by a purely conventional or quadratic network unless an exponential number of neurons are used \cite{liao2022heterogeneous}. Moreover, regarding the training cost, using the homogeneous type of neurons takes the extra learning cost to wire neurons specifically and orchestrate the learning process relative to using different neurons beforehand. In other words, involving neuronal diversity in artificial networks is an embodiment of modularization at the neuronal level. The development of modern industry has confirmed the superiority of modularization in efficiency.

Due to the spatiotemporal information processing ability, spiking neurons are highly energy-efficient. Unlike the conventional neuron that keeps the working status all the time, the spiking neuron idles unless it receives a spike from some events.

\textbf{Memory:} The reason why a connectionist model suffers catastrophic forgetting is that training such a model for a new task catastrophically interferes with the knowledge amassed in the previous task. In contrast, although humans tend to gradually forget previous information as getting old, learning new knowledge while catastrophically forgetting old knowledge rarely happens. Achieving lifelong learning is difficult because of the stability-plasticity dilemma: the model has to maintain both plasticity to acquire new knowledge and stability to prevent the consolidated knowledge from being dismantled. Roughly, three types of computational approaches have been proposed to address the catastrophic forgetting \cite{parisi2019continual}: i) imposing constraints on the level of plasticity to protect the consolidated knowledge \cite{li2017learning}; ii) allocating additional neural resources for new tasks \cite{rusu2016progressive}; iii) using two complementary learning systems dedicated to learning new information and replaying old experiences, respectively \cite{shin2017continual}. Inspired by the motivation in the second type of approaches, we find that polynomial neurons can offer a novel view to achieve lifelong learning by enabling a network to be internally progressive. Thus, knowledge for old tasks is stored instead of destroyed when facing new tasks. As shown in Figure \ref{fig:internally_progressive_network}(a), traditionally, given a new task, a new sub-network is created, and lateral links with the old tasks are learned. We refer to such a network as an externally progressive network. In contrast, assume a polynomial neuron is used, as Figure \ref{fig:internally_progressive_network}(b), tensor decomposition is doable to rearrange a polynomial neuron \cite{chrysos2021deep} into an internally compositional structure which encodes knowledge of a sequence of tasks into a hierarchy. Retaining old terms and adding new terms provide flexibility to balance old knowledge and the new. The internally progressive mechanism may apply to the situation where new tasks and old tasks are somewhat relevant. 

\textbf{Interpretability:} One way to derive an interpretation from a model is to understand its components, as the entire complex system can be usually decomposed into a combination of many functional modules \cite{lipton2018mythos}. For example, Bau et al. \cite{bau2020understanding} analyzed a CNN trained on the scene classification task and discovered via the receptive field analyses that each neuron matches certain object concepts. In the same vein, an exciting point about a polynomial neuron or other nonlinear neurons is that the neuron \textit{per se} contains an internal attention mechanism. The following derivation shows how to cast the attention mechanism from a quadratic neuron \cite{liao2022attention}:
\begin{equation}
\begin{aligned}
&\sigma((\boldsymbol{x}^\top\boldsymbol{w}^{r}+b^{r})(\boldsymbol{x}^\top\boldsymbol{w}^{g}+b^{g})+(\boldsymbol{x}\odot\boldsymbol{x})^\top\boldsymbol{w}^{b}+c) \\
=&\sigma (\boldsymbol{x}^{\top}\boldsymbol{w}^g\left( \boldsymbol{x}^{\top}\boldsymbol{w}^r+b^r \right) +b^g\boldsymbol{x}^{\top}\boldsymbol{w}^r+b^gb^r+\left( \boldsymbol{x}\odot \boldsymbol{x} \right) ^{\top}\boldsymbol{w}^b)\\
=&\sigma ( \boldsymbol{x}^{\top}\left( \boldsymbol{w}^g\left( \boldsymbol{x}^{\top}\boldsymbol{w}^r+b^r \right) \right) +\boldsymbol{x}^{\top}\left( \boldsymbol{w}^rb^g \right)
+\boldsymbol{x}^{\top}\left( \boldsymbol{x}\odot \boldsymbol{w}^b \right) ) \\
= & \sigma ( \boldsymbol{x}^{\top}( \underbrace{\boldsymbol{x}\odot \boldsymbol{w}^b+\boldsymbol{w}^g \boldsymbol{x}^{\top}\boldsymbol{w}^r}_{attention} +\underbrace{\boldsymbol{w}^g b^r+\boldsymbol{w}^rb^g}_{bias})) ,
\end{aligned}
\label{eq:quadratic}
\end{equation}
where $\boldsymbol{x}\odot \boldsymbol{w}^b+\boldsymbol{w}^g \boldsymbol{x}^{\top}\boldsymbol{w}^r$ can reflect where a neuron deems as important regions, in analogy to the attention mechanism \cite{vaswani2017attention}.
Furthermore, in this light, all neurons with a nonlinear aggregation function are self-explanatory. Suppose that a neuron is $\sigma(g(\x))$, we conduct the Taylor expansion around 0 for $g(\x)$ and remove the third and higher-order terms:
\begin{equation}
\begin{aligned}
       \sigma(g(\x)) & = \sigma(g(0)+\x^\top  D\mathbf{g}(0)+ \x^\top H\mathbf{g}(0) \x)\\
       &= \sigma(g(0) + \x^\top (\underbrace{D\mathbf{g}(0)+H\mathbf{g}(0) \x}_{attention})), 
\end{aligned}
\end{equation}
where $D\mathbf{g}(0)$ is the partial derivative of $g(\x)$ at zero, and $H\mathbf{g}(0)$ is the Hessian matrix of $g(\x)$ at zero. Clearly, the mainstream neuron type $\sigma((\boldsymbol{x}^\top\boldsymbol{w}+b)$ does not enjoy such a kind of self-interpretability, which necessitates the involvement of new type of neurons in artificial networks for better interpretability.

\section{Representative Applications}
Since introducing neuronal diversity is a fundamental change instead of a slight modification for a neural network, it has a global impact on the research and development of artificial networks, with the promise of pushing a wide spectrum of applications. Here, we discuss representative real-world applications in different important fields to illustrate the practical value of introducing neuronal diversity.

\subsection{Medical Imaging}

X-ray computed tomography (CT) is one of the most popular and important imaging modalities in hospitals and clinics. Although CT can offer critical clinical
information, patients have to bear potential risks because X-rays may cause genetic changes and cancer \cite{macmahon1962prenatal}. Therefore, reducing the radiation dose is an important problem in the CT field. However, images reconstructed from the lower dose suffer from noise and other kinds of artifacts. In clinics, noise removal, texture preservation, and
structure fidelity are three key aspects concerning radiologists. Algorithms should achieve a reasonable balance between these three aspects for a better clinical diagnosis.

Autoencoders \cite{vincent2010stacked} are a class of networks that consist of encoding and decoding parts. The encoding part attempts to learn a new representation, and the decoding part regenerates the input from the learned representation. In \cite{fan2019quadratic}, a quadratic autoencoder (Q-AE) was proposed to process the low-dose CT images, in hope that the quality of processed images can reach the level of images reconstructed from normal-dose CT. This Q-AE model employs ReLU as activation functions for all neurons, and has 5 quadratic convolutional layers in the encoding and 5 quadratic deconvolutional
layers in the decoding, where each layer has 15 quadratic
filters of $5\times5$, and symmetric layers are aggregated by residual connections. The anonymous reader study on the Mayo low-dose CT dataset revealed the superior performance of the quadratic autoencoder in terms of image denoising and model efficiency than other state-of-the-art models.

\subsection{Industrial Informatics}

The bearing faults are the most common source of faults in rotating machines such as wind turbines and aircraft engines \cite{bonnett2008increased}. To enhance the reliability of rotating machines and avoid economic loss, accurately diagnosing bearing faults is of great importance. Currently, a common and viable diagnosis method is to first measure vibration signals by attaching the measurement instrument to the rotating bearing \cite{mcfadden1984model}, and then use an artificial network to classify the faulty signals from the normal. Despite that artificial networks have achieved great successes in bearing fault detection, it lacks interpretability, \textit{i.e.}, it is hard to know if the model conforms to the physics principle when classifying fault signals out.

A convolutional neural network made of quadratic neurons (QCNN) was proposed for bearing fault diagnosis \cite{liao2022attention}. With the \textit{qttention} mechanism as derived in Eq. \ref{eq:quadratic}, the feature extraction process of QCNN and the physics principle explaining why the model can deliver good classification performance are deciphered to a large extent. For example, by visualizing faulty bearing signals and the qttention maps, it was found that all the faulty areas were captured by QCNN. By comparing the raw signal and the qttention map in the frequency domain, it was discovered that QCNN favors bearing fault frequency elements over the shaft frequency elements. 

\subsection{Numerical Computing}

A general form of a non-linear partial differential equation (PDE) can be expressed as $f(u,\gamma)=0$, where $f$ is a non-linear operator performed on partial
derivatives of the target variable $u$ (\textit{e.g.}, $\frac{\partial u}{\partial x}, \frac{\partial u}{\partial t}, \frac{\partial^2 u}{\partial x^2},$),
and $\gamma$ represents the parameters of the PDE. In the realm of PDEs and numerical computing, two classes of problems are mainly concerned: (a) forward problems, which are to solve for the target variable
$u$ prescribed by the PDE, and (b) inverse problems, which involve learning the unknown parameters, $\gamma$,
of the PDE, given the observations of $u$ at different timestamps.

Physics-informed neural networks (PINNs) are a type of neural networks that consider the physical laws and prior knowledge governing the problem in model design and training \cite{karniadakis2021physics}. The prior knowledge of general physical laws is in the form of PDEs, and can be employed as a regularization for the training of a network, \textit{e.g.}, formulating the PDE equation into a supervised loss function. Motivated by the nonlinear approximation ability of quadratic networks, Bu \textit{et al.} \cite{bu2021quadratic} proposed a quadratic residual network to solve the forward and inverse problems in PDEs. Following the original PINN framework, empirical results demonstrated that QResNet exhibits consistent advantages over conventional networks in terms of parameter efficiency and approximation accuracy. Let us take the Allen-Cahn equation \cite{shen2010numerical} as an example to show why QResNet fits. The task is to predict $u$ based on the PDE as follows:
\begin{equation}
    \frac{\partial u}{\partial t} -0.0001\cdot\frac{\partial^2 u}{\partial x^2} +5u^3 -5u=0 
    \label{eqn:allen_cahn}
\end{equation}
and the initial condition. Discretizing this equation leads to 
\begin{equation}
\underbrace{u_{t+1}=u_t}_{residual}+\underbrace{0.0001\cdot\frac{\partial^2 u_t}{\partial x^2} -5u_t^3 +5u_t}_{polynomial ~relation}. 
\label{eq:residual}
\end{equation}
Since the right-hand side of Eq. \ref{eq:residual} consists of a polynomial function regarding $u_t$. Therefore, it is more suitable to use a quadratic neuron to learn it than a conventional neuron. Moreover, Eq. \ref{eq:residual} contains a residual relation, which well fits a residual network.

\subsection{Computer Vision}

Computer vision is an interdisciplinary field that enables a computer to derive a meaningful understanding from digital images, videos, or other visual inputs \cite{szeliski2022computer}. Computer vision tasks concern acquiring, processing, analyzing, and understanding digital inputs such as image restoration, face recognition, and video tracking.
Neural network models have been dominating computer vision since the tremendous success in classifying approximately 1.2 million images into 1000 classes in the ImageNet challenge \cite{krizhevsky2012imagenet}. 

As a drop-in replacement, polynomial networks \cite{chrysos2021deep} have been applied to a plethora of computer vision tasks. It was empirically demonstrated that the polynomial networks produce competitive results in a large variety of computer vision tasks such as image recognition, face recognition, and image generation.

\section{Challenges and Outlooks}


\begin{figure}
\centering
\includegraphics[width=\linewidth]{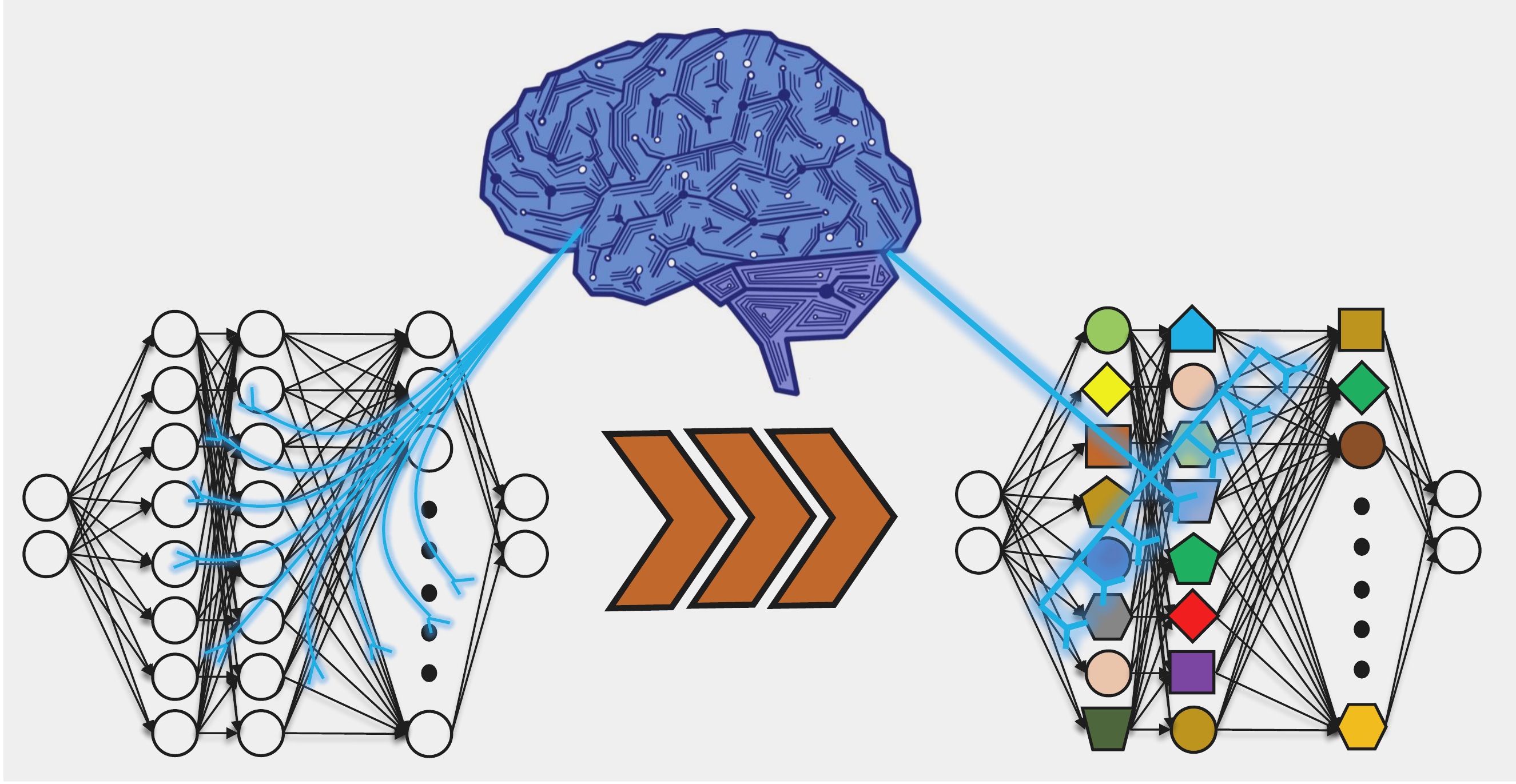}
\caption{Different neurons should be synergized together to maximize their strengths. Despite the diversity, neurons have several basic types (denoted by various shapes), and each type of neurons have similar but slightly different variants (denoted by various colors).} 
\label{fig:neuronal_synergy}
\end{figure}

\textbf{Neuronal synergy.} To unleash the potential of neuronal diversity, the core problem is how to make different neurons synergize together to maximize their strengths, as Figure \ref{fig:neuronal_synergy} shows. In the human brain, the activities of a large number of neurons are well coordinated. For example, parallel activities of neurons are observed to lie in a low-dimensional manifold \cite{gallego2017neural,gallego2018cortical}. Although the underlying coordination mechanism remains unclear, we know that the coordination of different neurons in the brain is highly artful and efficient, \textit{e.g.}, the coordination of distant neurons is enabled not only via long-range connections but also through heterogeneity in local connectivity \cite{dahmen2022global}. Generally speaking, the learning of an artificial network is governed by the loss function and the optimization algorithm. There is no explicit algorithm suggesting how to synergize different types of neurons for the same task. Inspired by neural architecture search \cite{elsken2019neural}, a brute-force means to accommodate this problem is neuronal cell search, which takes the neuronal type as the model's hyperparameters to optimize in the framework of autoML \cite{he2021automl}. However, it is more desirable if neuroscience findings can shed light on some rule-of-thumb or useful inductive bias to guide the neuronal synergy. For example, in the population coding theory, the collective responses of a population of neurons are to maximize the amount of information \cite{tkavcik2010optimal}.

\textbf{Task-based neuron design.} The past 10 years have witnessed a surge of many outstanding architecture designs, such as U-Net \cite{ronneberger2015u}, the pyramidal structure \cite{han2017deep}, and shortcuts \cite{he2016deep,fan2021sparse}. The central principle behind these studies is designing a network architecture according to the needs of a task. In light of neuronal diversity, the neuronal type is also critical to the power of a neural network. Thus, we ask the following question: \textit{Can the network design go from the task-based architecture design to the task-based neuron design?} Our brain is exactly a task-based neuron designer, \textit{i.e.}, biological neurons have abundant functional diversity, which is a necessity for the brain to execute different tasks. The advantage of the task-based neuron design over the task-based architecture design is that task-specific neurons contain useful implicit bias for the task. Thus, the network of these task-specific neurons can integrate the task-driven forces of all these neurons, which should be much stronger than the network of generic neurons with the same structure. Our conjecture is that the task-based neuron design will escalate neural network research into a new stage and make advances in many previously-believed difficult tasks. Figure \ref{fig:task_based_neuron_design} showcases a three-step roadmap for the task-based neuron design:

\begin{figure}
\centering
\includegraphics[width=\linewidth]{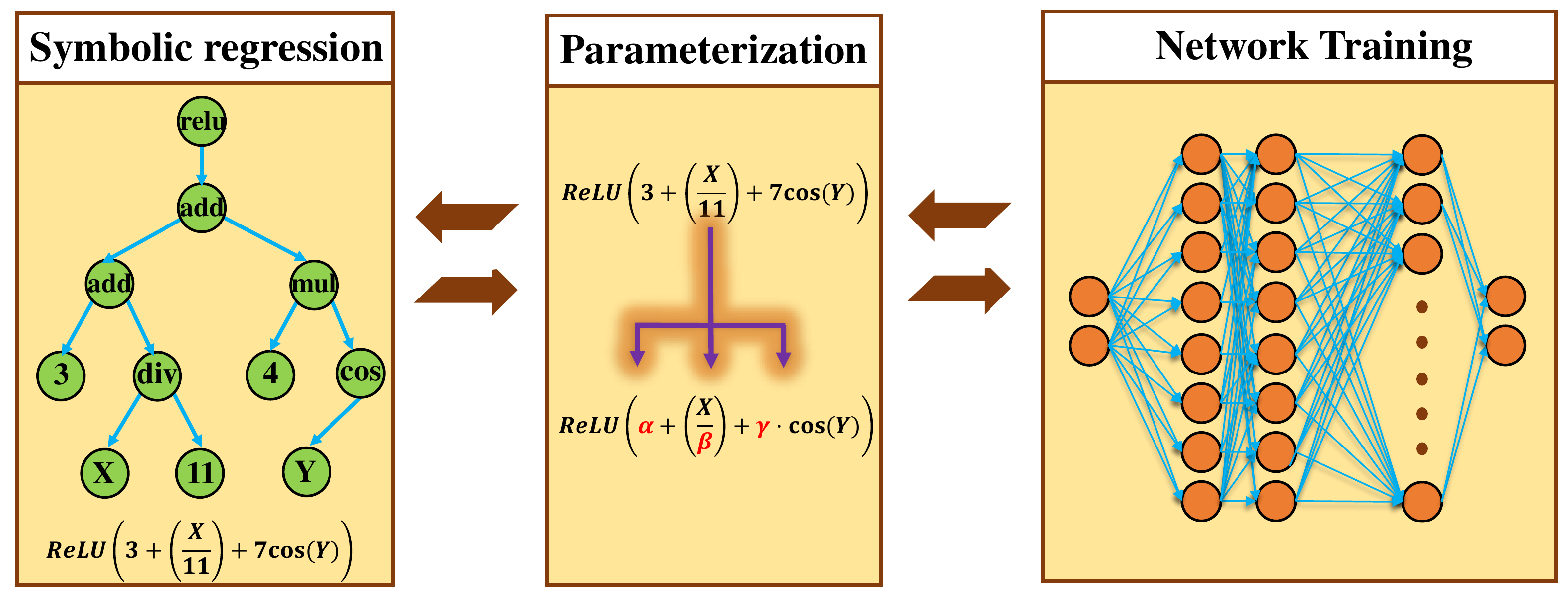}
\caption{A roadmap for designing and deploying task-based neurons: i) build an elementary neuronal model by the symbolic regression; ii) parameterize the acquired elementary neuron to make its parameters learnable; iii) employ task-based neurons in a network for validation and feedback. } 
\label{fig:task_based_neuron_design}
\end{figure}

$\bullet$ Build an elementary neuronal model via symbolic regression. Encouraged by the concept of linear regression or polynomial regression, symbolic regression \cite{schmidt2009distilling} is to search over the space of all possible mathematical formulas regarding the input variables, starting from base functions such as logarithmic, trigonometric, and exponential functions. Furthermore, the search space should be regularized to entitle the established neuron with the desirable properties, \textit{e.g.}, no gradient vanishment or explosion when connected to a network. For example, we can split the search space into two parts: the aggregation function and the activation function. 

$\bullet$ Parameterize the learned elementary neuron to make its parameters trainable. Such parameterization can be straightforwardly made by casting all coefficients in the elementary neuron as learnable parameters. However, for better efficiency and expressivity, selecting which coefficients to parameterize can be optimized based on the final performance. 

$\bullet$ Employ task-based neurons in a network for validation and feedback. For example, according to the performance of the network, some base formulas in a task-based neuron might be pruned. Task-based neurons can also facilitate multi-task learning. With increasingly many task-based neurons designed, a warehouse of neurons can be gradually established in which a knowledge graph and neuron-based informatics could be developed.

\textbf{Theoretical issues.} Since the rise in 2012, deep learning has been criticized for lacking a fundamental theory. This embarrassment is greatly alleviated in recent years with the rapid development of deep learning theory. Expressivity-wise, why a deep network performs superbly \cite{cohen2016expressive,poole2016exponential} is well addressed by characterizing the complexity of a function expressed by a neural network, \textit{i.e.}, increasing depth can greatly maximize such a complexity measure compared to increasing width; the power of shortcuts are demonstrated as well \cite{fan2021sparse}, \textit{i.e.}, a network with shortcuts can express a far more complicated network than a network without shortcuts. Optimization-wise, the neural tangent kernel theory suggests that the training of an infinitely wide network is equivalent to a kernel ridge regression \cite{jacot2018neural}. The understanding of the generalization ability of deep networks is also deepened by the discovery of the double descent phenomenon \cite{belkin2019reconciling}. 


Notwithstanding, the existing theory provides an explanation for the simplest neuron that is based on an inner product and an activation function, which may not be applicable to other kinds of nonlinear neurons. In the context of neuronal diversity, we ask the following question for theorists to brainstorm: \textit{to what extent are the current theories scalable to the heterogeneous networks?} Addressing this question is highly nontrivial in two senses. On the one hand, the philosophy behind homogeneous and heterogeneous networks varies greatly. The former implicitly assumes that a universal type of neurons can solve a wide class of complicated nonlinear problems, simply referred to as "one-for-all". Such a philosophy is well supported by the universal approximation theorem \cite{hornik1990universal}. However, the problem of this philosophy is practicality and efficiency, \textit{i.e.}, it may suffer the curse of dimensionality. In contrast, the latter assumes different types of neurons solve a specific problem, which is referred to as "all-for-one". The loss of universality in heterogeneous networks adds a layer of complication to the theoretical analyses. The low-hanging fruits may come from the efficiency analysis first by showing that the mode of "all-for-one" is more efficient than that of "one-for-all". Then, the optimization and generalization properties can be further analyzed.

\textbf{Neuroinformatics of artificial neurons and networks.} Neuroinformatics is an interdisciplinary field that introduces informatics into neuroscience for data mining and information processing. With the large-scale deployment of those task-specific neurons into networks, methodologically, we think it is highly necessary to introduce tools of informatics into artificial neurons and networks to further extract information from different neurons and networks, referred to as neuroinformatics of artificial neurons. The goals of this kind of neuroinformatics are to make contributions to the next generation of AI by drawing insights into the information processing of biological networks, supporting brain-inspired intelligence, and fertilizing the interpretability of artificial networks. To realize this goal, on the one hand, tools for analyzing and standardizing artificial neurons and networks should be developed; on the other hand, the database and knowledge graph can be built for various artificial neurons to supply ontology information about neuron fitness and relations between different neurons, to advance the connectivity organization, and to instruct the multi-modules synergy. We believe that the establishment of neuroinformatics of artificial neurons will promote NeuroAI to a higher level.

\begin{figure}
\centering
\includegraphics[width=\linewidth]{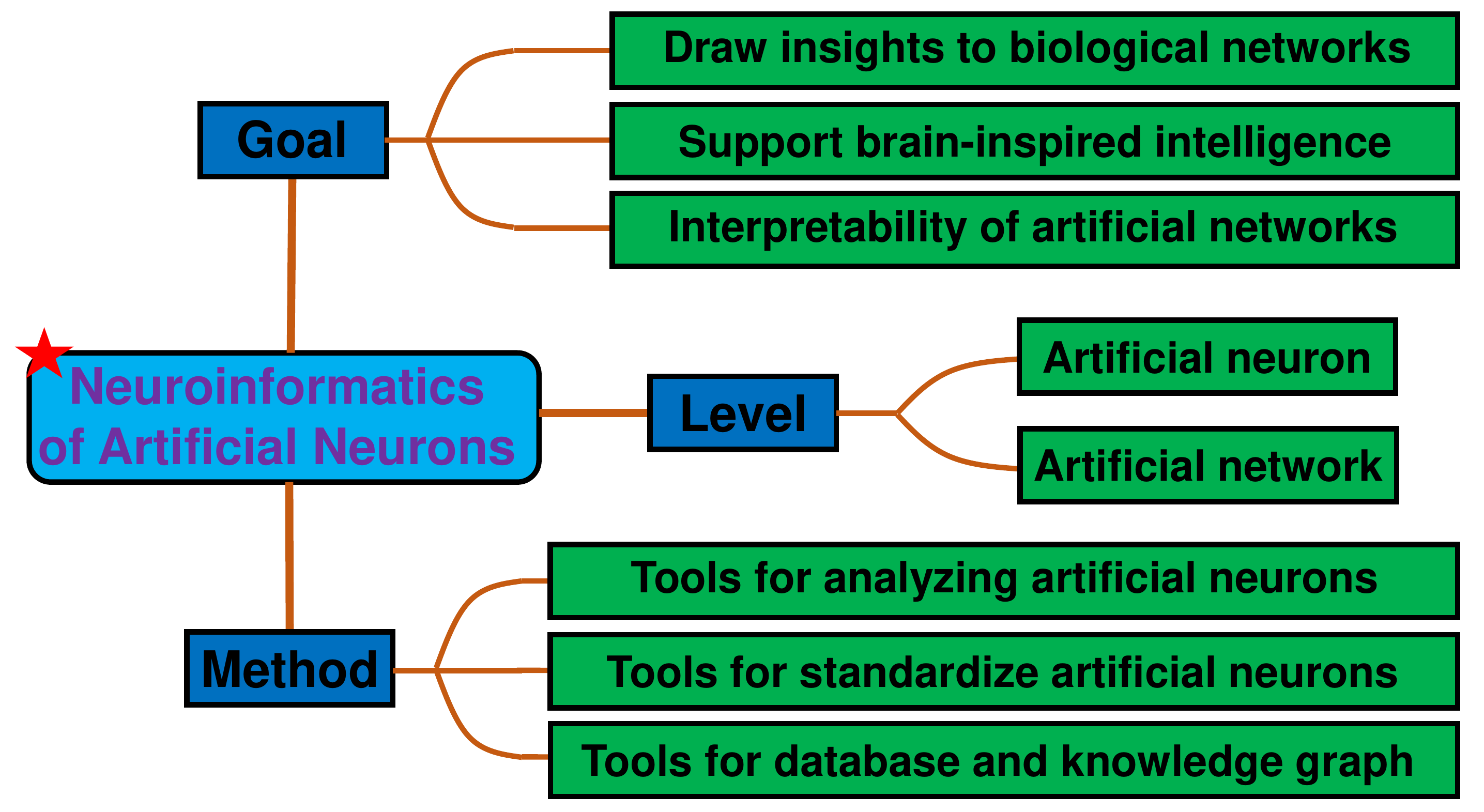}
\caption{The goal and methods of neuroinformatics of artificial neurons and artificial networks.}
\label{fig:neuronal_neuroinformatics}
\vspace{-0.5cm}
\end{figure}

\section{Resource Summary}
We list the useful resources about neuronal diversity in Table \ref{tab:resouce_list} for readers' reference.

\begin{table*}
\caption{A Resource Summary for Neuronal Diversity in Artificial Networks. }
\centering
\scalebox{1.0}{
\begin{tabular}{c|c|c}
\hline
    \hline
    \rowcolor{LightCyan}
    \textbf{Resource} & \textbf{Type} & \textbf{Description} \\ 
    \hline
    \href{https://proceedings.mlsys.org/paper/2022/hash/a5bfc9e07964f8dddeb95fc584cd965d-Abstract.html}{QuadraLib}   &  Library & \makecell{The QuadraLib is a library for the efficient optimization and design exploration of quadratic networks. \\
    The paper of QuadraLib won MLSys2022's best paper award.}  \\  
    \hline 
    \href{https://github.com/FengleiFan}{Dr. Fenglei Fan's GitHub Page}   & Code & \makecell{Dr. Fenglei Fan's GitHub Page summarizes a series of papers and associated code on quadratic networks, \\
    including quadratic autoencoder and the training algorithm \textit{ReLinear}.}  \\ 
    \hline 
\href{https://github.com/grigorisg9gr/polynomial_nets}{Polynomial Network}   & Code & This repertoire shows how to build a deep polynomial network and sparsify it with tensor decomposition. \\    \hline 
    \href{http://www.dendrites.org/dendrites-book}{Dendrite}   & Book & A comprehensive book covering all aspects of dendritic computation. \\   
    \hline
\end{tabular}}
\label{tab:resouce_list}
\end{table*}

\section{Conclusions}

In this perspective, we have systematically introduced neuronal diversity into artificial networks, as a practice of NeuroAI, including biological background, the existing studies, challenges, and future directions. We believe that neuronal diversity has the potential of elevating artificial networks into a new age. Future efforts can be invested to demonstrate more and more killer applications of new neurons. 

\section*{Acknowledgements}

The author is grateful for the inspiring discussions with Mr. Hang-Cheng Dong (HIT) and Mr. Jing-Xiao Liao (HIT).

\bibliographystyle{unsrt}
\bibliography{example}

\end{document}